%% file: main.tex
\documentclass{article}

\usepackage{PRIMEarxiv}

\usepackage[utf8]{inputenc} 
\usepackage[T1]{fontenc}    
\usepackage{url}            
\usepackage{booktabs}       
\usepackage{amsfonts}       
\usepackage{nicefrac}       
\usepackage{microtype}      
\usepackage{fancyhdr}       
\usepackage{graphicx}       
\usepackage{amsmath}
\usepackage{amssymb}
\usepackage{amsthm}
\usepackage{float}          
\usepackage{enumitem}       
\usepackage[numbers,sort&compress]{natbib}
\usepackage[dvipsnames]{xcolor}
\usepackage[colorlinks=true,linkcolor=MidnightBlue,citecolor=MidnightBlue,urlcolor=MidnightBlue]{hyperref}

\newtheorem{remark}{Remark}
\newcommand{\fit}{\textsc{fit}}
\newcommand{\cala}{\textsc{cal}\textsubscript{A}}
\newcommand{\calb}{\textsc{cal}\textsubscript{B}}
\newcommand{\eval}{\textsc{eval}}
\input{generated/numbers}

\title{Abstention Spends Its Budget on Local Label Variance More Than on Novelty:\\[6pt]
{\large Auditing Conformal Calibration and Selective Prediction under Molecular Scaffold Shift\par}
}

\author{
  Arnav Srivastava \\
  New York University \\
  \texttt{as19397@nyu.edu}
  \And
  Krishna Kassetty \\
  PhonePe \\
  \texttt{kassettykrishna@gmail.com}
}

\begin{document}
\maketitle

\begin{abstract}
Selective prediction with split conformal calibration is an attractive post-hoc fix for the translation gap in
molecular property prediction: abstain on the least reliable compounds, and the retained set should become
trustworthy without retraining. The implicit hope is that abstention concentrates on out-of-distribution
chemistry. We audit this recipe for a gradient-boosted fingerprint model on one regression (lipophilicity) and one
imbalanced classification (BBBP) endpoint under a Bemis--Murcko scaffold split, holding the fitted model and the
evaluation molecules fixed while varying only the calibration fold, over \NRep{} replicates. (i)~On lipophilicity,
whether conformal calibration restores coverage depends on where the calibration fold comes from: calibrating on
held-out molecules from the training scaffolds falls below nominal (PICP \LipoPicpA{} at a target of $0.90$), while
calibrating on novel-scaffold molecules recovers it (\LipoPicpB); on BBBP, coverage stays above nominal in every
regime. (ii)~For split conformalized quantile regression the conformal
correction is a constant, so ranking by interval width is ranking by the base model's quantile width;
conformalization does not change \emph{what} is abstained on. (iii)~The abstention score (CQR width on
lipophilicity; effectively ensemble variance on BBBP) tracks local label variance more than chemical novelty:
controlling for local label variance largely explains away the novelty association (partial Spearman \LipoPartNov{}
and \BbbpPartNov). (iv)~At the pre-specified operating point and across an exploratory sweep of \GateWidthCells{}
budget--threshold settings, gating on interval width or ensemble variance never improves hit precision over ranking by
point prediction; a plain novelty gate shows small, uncorrected gains at the smallest budgets. (v)~In an exploratory
analysis, exact conformal selection with
Benjamini--Hochberg exceeds its false discovery rate target at every level under training-scaffold calibration
(FDR \JcATwentyFdr{} at $q{=}0.2$) and is statistically consistent with it under novel-scaffold calibration; its
contribution is a data-driven batch size with an error guarantee rather than a better ranking. In these
experiments, conformalization earns its keep through guarantees whose validity hinges on calibration provenance,
not by choosing what to abstain on.
\end{abstract}


\section{Introduction}
In design--make--test--learn loops, a property predictor that is accurate on a random split but silently
overconfident on a new chemical series spends scarce assay budget on wrong predictions it presents as certain.
Uncertainty quantification is the standard response \citep{hirschfeld2020uncertainty,scalia2020evaluating},
and split conformal prediction \citep{vovk2005algorithmic,papadopoulos2002inductive,lei2018distribution,
angelopoulos2023gentle} is especially attractive: it wraps any fitted model, requires no retraining, and
delivers a finite-sample, distribution-free coverage guarantee under exchangeability. Paired with selective
(reject-option) prediction \citep{elyaniv2010foundations,geifman2017selective}, it promises a decision-aware
operating point: abstain on the least reliable fraction and the retained set becomes trustworthy. The
implicit hope is that abstention concentrates on \emph{out-of-distribution} chemistry---novel scaffolds a model
should not be confident about.

We audit that hope in a controlled setting. On the \emph{same} fitted model and the \emph{same} evaluation
molecules, we contrast calibration on held-out molecules from the training scaffolds with calibration on molecules
from novel scaffolds, which is exchangeable with the evaluation set. We then ask what the abstention signal
selects, whether abstaining improves fixed-budget hit finding, and where conformal guarantees do pay off. We find:
\begin{enumerate}[label=(\roman*)]
\item \textbf{Coverage depends on calibration provenance.} On lipophilicity, training-scaffold calibration falls below
nominal under scaffold shift, while novel-scaffold calibration recovers it, on identical models and molecules; on BBBP,
coverage stays above nominal in every regime (\S\ref{sec:e1}).
\item \textbf{Conformalization does not change the abstention ranking.} For split CQR the calibrated interval is the
base quantile interval widened by a constant (Remark~\ref{rem:width}), so width-based abstention inherits the base
model's ranking (\S\ref{sec:e2}).
\item \textbf{The ranking tracks local label variance more than novelty.} Its association with novelty is largely
explained away by local label variance, a quantity that mixes measurement noise with structure--activity
heterogeneity and itself rises with novelty (\S\ref{sec:e2}).
\item \textbf{Uncertainty gates do not beat point-prediction ranking at fixed budget.} At the pre-specified operating
point and across an exploratory sweep, width and ensemble-variance gates are never better and usually significantly
worse; a novelty gate is better only at the smallest budgets (\S\ref{sec:e3}).
\item \textbf{FDR control in conformal selection depends on calibration provenance} (exploratory). Exact conformal
selection \citep{jin2023selection} exceeds its target at every level under training-scaffold calibration and is
consistent with it under novel-scaffold calibration (\S\ref{sec:fdr}).
\end{enumerate}
Findings (i) and (v) confirm, in a controlled comparison, what exchangeability predicts and what drift studies on
toxicity data have reported \citep{morger2022drift}; findings (ii)--(iv) concern what abstention actually selects.
The scope is one model family, two endpoints and one scaffold split (\S\ref{sec:disc}).

\section{Related work}
\textbf{Conformal prediction and shift.} Split (inductive) conformal prediction
\citep{vovk2005algorithmic,papadopoulos2002inductive,lei2018distribution} gives marginal coverage under
exchangeability; conformalized quantile regression (CQR) adapts interval width to the input
\citep{romano2019cqr}, and adaptive prediction sets (APS/RAPS) do so for classification
\citep{romano2020classification,angelopoulos2021uncertainty}; class-conditional (Mondrian) calibration targets
per-class coverage \citep{vovk2012conditional}. Under covariate shift, weighted conformal prediction reweights
calibration scores by a density ratio \citep{tibshirani2019conformal}, and coverage gaps can be bounded beyond
exchangeability \citep{barber2023beyond}. In molecular settings, conformal prediction has been proposed as an
applicability-domain alternative \citep{norinder2014introducing}, adapted to feedback covariate shift in design
\citep{fannjiang2022conformal}, combined with density estimation for drug properties \citep{laghuvarapu2023codrug},
and shown to lose validity under temporal drift unless the calibration set is updated with data closer to the test
set \citep{morger2022drift}.

\textbf{Uncertainty, calibration and selective prediction.} Selective prediction trades coverage for risk via a
reject option \citep{elyaniv2010foundations,geifman2017selective}. Ensembles are a strong uncertainty baseline
\citep{lakshminarayanan2017simple}; uncertainty is often decomposed into aleatoric and epistemic parts
\citep{kendall2017uncertainties}; calibration is typically measured with ECE \citep{guo2017calibration,
nixon2019measuring} and degrades under dataset shift \citep{ovadia2019trust}. Molecular UQ benchmarks
\citep{hirschfeld2020uncertainty,scalia2020evaluating} find no uniformly dominant method, and similarity to the
training set has long been known to predict QSAR accuracy \citep{sheridan2004similarity}.

\textbf{Selection with error control.} Conformal selection \citep{jin2023selection} turns conformal $p$-values
into a candidate list with Benjamini--Hochberg \citep{benjamini1995controlling} false discovery rate (FDR)
control, which matches the assay-batch decision more directly than marginal coverage; weighted conformal
$p$-values extend it to covariate shift \citep{jin2026weighted}.

\textbf{Molecular shift and label heterogeneity.} Scaffold splits \citep{bemis1996properties,wu2018moleculenet}
and time splits \citep{sheridan2013time} emulate prospective deployment. Activity cliffs---structurally similar
molecules with very different properties \citep{stumpfe2012activity}---are a known failure mode of molecular ML
\citep{vantilborg2022exposing} and one source of the local label heterogeneity we measure.

\section{Setup}
\label{sec:setup}
\textbf{Data and model.} Lipophilicity ($n{=}\NLipo$, regression, label s.d.\ \LipoSd) and BBBP ($n{=}\NBbbp$,
classification, positive rate \BbbpPos) from MoleculeNet \citep{wu2018moleculenet}, canonicalized and deduplicated
by SMILES. Features are 2048-bit ECFP4 \citep{rogers2010extended} plus ten RDKit \citep{landrum_rdkit} descriptors.
The base predictor is LightGBM \citep{ke2017lightgbm}: a mean model, two quantile models at levels $\alpha/2$ and
$1-\alpha/2$ for CQR, and three bootstrap-bagged members for an ensemble-variance baseline.

\textbf{Split and calibration regimes.} A Bemis--Murcko scaffold split \citep{bemis1996properties} assigns scaffold
groups, largest first (ties, including all singleton scaffolds, broken by row order in the deduplicated dataset), to a
pool holding $70\%$ of the molecules; this split is deterministic and identical in every
replicate. Because the pool absorbs every multi-molecule scaffold group, all remaining molecules (\LipoNRem{} for
lipophilicity, \BbbpNRem{} for BBBP) have scaffolds that occur only once in the dataset; this is our notion of
novel chemistry. In each replicate a random $15\%$ of the pool is held out as \cala, which shares scaffolds with
training, and the model is fit on the rest (\fit); the novel remainder is split uniformly at random into \calb{}
($34\%$) and \eval{} (lipophilicity: $|\fit|{=}\SzFit$, $|\cala|{=}\SzCalA$, $|\calb|{=}\LipoNCalB$,
$|\eval|{=}\LipoNEval$; \OverlapFitEval{} \fit--\eval{} scaffold overlap). Regime~\textbf{A} calibrates on \cala,
which is not exchangeable with \eval. Regime~\textbf{B} calibrates on \calb, which is exchangeable with \eval{}
given the remainder, so split-conformal coverage and the conformal-selection FDR guarantee hold exactly in~B; it is
an oracle in that it requires labels on novel-scaffold molecules. A random split with the same block sizes is an
in-distribution control. Replicates thus re-draw \cala, the \calb/\eval{} assignment and the model seeds within one
fixed scaffold split (mean Jaccard overlap of \eval{} sets across replicate pairs: \LipoEvalJaccard{} for
lipophilicity), so all intervals are conditional on that split.

\textbf{Conformal procedures.} We use split CQR \citep{romano2019cqr} for regression and randomized APS
\citep{romano2020classification} for classification at level $\lceil (n{+}1)(1-\alpha)\rceil/n$ with
$\alpha{=}0.10$. For CQR the calibration scores are $s_i=\max\{\hat q_{\mathrm{lo}}(x_i)-y_i,\;
y_i-\hat q_{\mathrm{hi}}(x_i)\}$, $\hat Q$ is their $\lceil (n{+}1)(1-\alpha)\rceil$-th smallest value, and
$C(x)=[\hat q_{\mathrm{lo}}(x)-\hat Q,\;\hat q_{\mathrm{hi}}(x)+\hat Q]$. Our APS implementation replaces an empty
prediction set with the top class, which adds coverage (Appendix~\ref{app:aps}).

\begin{remark}[Width ordering under split CQR]\label{rem:width}
$|C(x)| = \big(\hat q_{\mathrm{hi}}(x)-\hat q_{\mathrm{lo}}(x)\big) + 2\hat Q$, and $\hat Q$ does not depend
on $x$. Hence any rule that ranks or thresholds test molecules by $|C(x)|$ at a quantile of the test widths
selects exactly the same molecules as the same rule applied to the uncalibrated quantile width
$\hat q_{\mathrm{hi}}(x)-\hat q_{\mathrm{lo}}(x)$. The conformal step changes coverage, not the ordering.
\end{remark}
The same holds for our covariate-shift-weighted variant, which reweights calibration scores by a
domain-classifier density ratio \citep{tibshirani2019conformal} but replaces the test-point weight with the
largest calibration weight (a heuristic, not a bound), so that $\hat Q$ remains a scalar; the exact construction of
\citet{tibshirani2019conformal} has an $x$-dependent $\hat Q(x)$ and could re-rank. For BBBP, prediction sets
contain one or two labels and \BbbpSizeOne{} of \eval{} molecules receive a single label in regime~A, so the
abstention score---set size with ties broken by ensemble variance---is effectively the bagged-ensemble variance
(Spearman \BbbpRhoWidthEns{} with it).

\textbf{Covariates and metrics.} For each \eval{} molecule we record \emph{molecular novelty} $1-\max$ Tanimoto
(ECFP4) to any \fit{} molecule; \emph{scaffold novelty}, the same on Murcko scaffolds; \emph{local label variance},
the (population) variance of labels among its $k{=}10$ highest-Tanimoto \fit{} neighbours; and the realized error
($|y-\hat y|$, or misclassification for BBBP). The enrichment of an abstained set for a covariate is the fraction
of abstained molecules in the covariate's top quartile divided by $0.25$; for BBBP errors it is the abstained
misclassification rate divided by the overall rate. For regression we report PICP and MPIW (empirical coverage and
mean width of the intervals) and RMSE; for classification, coverage, ECE (ten equal-mass bins), Brier score and NLL.

\textbf{Decision protocol.} A hit is a label at or above $c$, the $80$th percentile of \fit{} labels
(\HitTieFrac{} of \eval{} labels equal $c$). With an assay budget $B$ (a fraction of \eval), a gate removes the top
$(1-\kappa)$ fraction of \eval{} by its score (width, ensemble variance or novelty) and the remaining molecules are
ranked by point prediction $\hat\mu$ and the top $B$ selected; \textsc{naive} ranks all molecules by $\hat\mu$.
Conformal selection \citep{jin2023selection} tests the null $H_{0,j}\colon y_j<c$ with the clipped score
$V(x,y)=M\,\mathbf{1}\{y\ge c\}+c-\hat\mu(x)$ ($M$ exceeding the label range) and the randomized $p$-values
\begin{equation}\label{eq:jc}
p_j=\frac{N_j^{>}+U_j\,(1+N_j^{=})}{|\textsc{cal}|+1},\qquad U_j\sim\mathrm{Unif}(0,1),
\end{equation}
where $N_j^{>}$ and $N_j^{=}$ count calibration nulls ($y_i<c$) with $\hat\mu(x_i)>\hat\mu(x_j)$ and
$\hat\mu(x_i)=\hat\mu(x_j)$, respectively,
and then applies Benjamini--Hochberg at level $q$; this controls FDR at $q$ when calibration and test points are
exchangeable \citep{jin2023selection}. The count $N_j^{>}$ is monotone in $\hat\mu(x_j)$, so the selected set
consists largely of the top-$k$ molecules by point prediction for a data-dependent $k$, with the randomization
reordering molecules whose counts tie. We estimate FDR as the replicate
mean of the false discovery proportion, with FDP $:=0$ for an empty selection (the quantity BH controls).

\textbf{Statistics and pre-specification.} The primary analyses ($\alpha{=}0.10$; \NRep{} replicates; split
fractions; $k{=}10$; abstention budgets $\{5,10,20,30\}\%$ with top-quartile thresholds; assay budget $15\%$; hit
quantile $0.80$; FDR level $0.2$) were fixed in the configuration of the original \NRep-replicate experiments (SHA-256
prefix \ConfigHash), which predates all additional analyses. The abstention fraction $\kappa{=}0.8$ and the model hyperparameters were fixed in the original code but
are not part of the hash. The additional analyses---exact conformal selection, the FDR levels $0.1$ and $0.3$, the
budget--threshold sweep, the novelty gate, the uncalibrated-width control and the diagnostics in
Appendices~\ref{app:aps} and~\ref{app:noise}---were added later and are exploratory. Intervals are $95\%$ percentile-bootstrap intervals over
replicates (\NRepSweep{} replicates for the $\alpha$ sweep), policy comparisons are paired by replicate, and none is
corrected for multiple comparisons.

\section{Results}

\subsection{Coverage under shift depends on calibration provenance}
\label{sec:e1}
\begin{figure}[t]
\centering
\includegraphics[width=\linewidth]{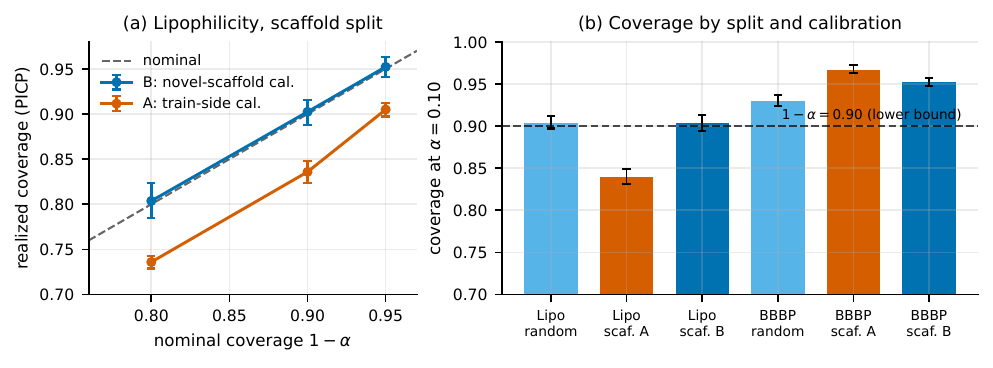}
\caption{\textbf{Coverage depends on calibration provenance.} (a)~Realized vs.\ nominal coverage on lipophilicity
under the scaffold split (\NRepSweep{} replicates; the base quantile models are fixed at levels $0.05/0.95$ and only
the conformal level varies): training-scaffold calibration (A) falls below nominal at every level, novel-scaffold
calibration (B) tracks it. (b)~Coverage at $\alpha{=}0.10$ by split and regime (95\% bootstrap intervals); BBBP values
include the non-empty-set rule.}
\label{fig:cal}
\end{figure}

On lipophilicity, calibration on training-scaffold molecules (A) falls below nominal: PICP \LipoPicpACI{} against a
target of $0.90$. Calibration on novel-scaffold molecules (B) is at nominal, \LipoPicpBCI, on the same model and
molecules, and so is the random-split control (\LipoPicpRandCI; Table~\ref{tab:e1}). The coverage guarantee does not
apply in~A, where calibration and test points are not exchangeable; varying only the calibration fold attributes the
shortfall to it. The gap persists across the $\alpha$ sweep (Figure~\ref{fig:cal}a): at nominal $0.95/0.90/0.80$,
regime~A attains \SweepAFive/\SweepATen/\SweepATwenty{} and regime~B \SweepBFive/\SweepBTen/\SweepBTwenty. (A Gaussian
miscalibration-area summary ranks A ahead of B; Appendix~\ref{app:miscal} explains why we do not use it.)

BBBP behaves differently: all three BBBP settings exceed $0.90$ (random \BbbpCovRandCI, A \BbbpCovACI, B \BbbpCovBCI).
In the exchangeable settings this is almost entirely the non-empty-set rule (it accounts for \RuleShareRand{} and
\RuleShareB{} of the excess; unforced coverage \UnfRandCov{} and \UnfBCov). In~A the rule accounts for only
\RuleShareA{} of the excess and unforced coverage is still \UnfACov{} (Appendix~\ref{app:aps}), an empirical outcome
without a guarantee. ECE and
the proper scores also improve under the scaffold split (ECE \BbbpEceRand$\to$\BbbpEceA; Brier
\BbbpBrierRand$\to$\BbbpBrierA; NLL \BbbpNllRand$\to$\BbbpNllA), alongside a shift in class balance (\eval{} positive
rate \EvPosRand{} in the random split vs.\ \EvPosA{} under the scaffold split). Marginal APS under-covers the minority
(non-penetrant) class even in the exchangeable random control (\BbbpCovMinRand), so the under-coverage itself is a
property of marginal APS; its size follows overall coverage and therefore the calibration fold (\BbbpCovMinB{} in~B,
\BbbpCovMinA{} in~A; paired difference \MinDiffABCI). Class-conditional calibration raises B's minority coverage to \MonBMondCovMin{} at the cost of larger sets
(\MonBMargSet{} $\to$ \MonBMondSet; Appendix~\ref{app:e4}).

\begin{table}[t]
\centering
\caption{\textbf{Coverage and calibration under scaffold shift} ($\alpha{=}0.10$; mean over \NRep{} replicates, 95\%
bootstrap interval). Lipophilicity falls below the $0.90$ target only under training-scaffold calibration (A). BBBP
coverage is inflated by the non-empty-set rule of our APS implementation (unforced values in
Appendix~\ref{app:aps}). Coverage depends on the calibration regime; ECE, Brier and NLL are properties of the fitted
model and are shared across regimes.}
\label{tab:e1}
\small
\input{generated/tab_e1}
\end{table}

\subsection{What does the abstention signal select? Local label variance more than novelty}
\label{sec:e2}
By Remark~\ref{rem:width}, the lipophilicity abstention score is the base model's quantile width: abstaining on the top
$20\%$ by conformal width and by uncalibrated quantile width selects identical batches in every replicate (maximum
absolute precision difference \WidthIdentityMaxAbs). For BBBP the score is effectively ensemble variance
(\S\ref{sec:setup}).

\begin{figure}[t]
\centering
\includegraphics[width=\linewidth]{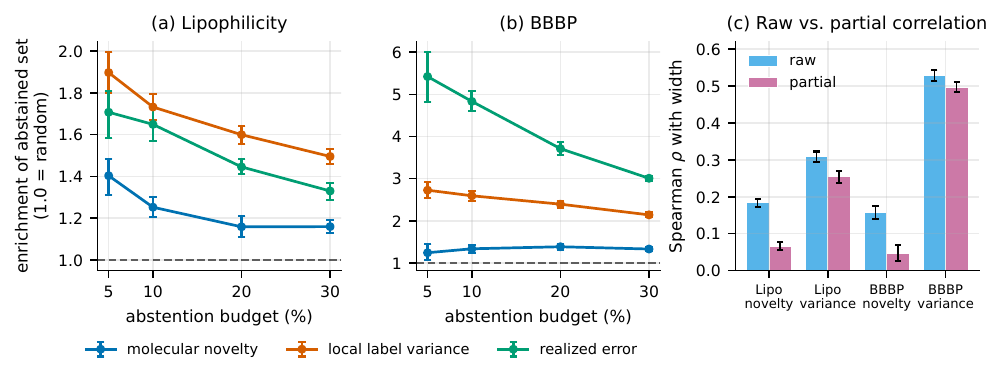}
\caption{\textbf{The abstention score follows local label variance more than novelty} (regime~A). (a,b)~Enrichment of
the abstained set relative to random abstention (1.0) for molecular novelty, local label variance and realized error
across abstention budgets; on BBBP (b) the score is effectively ensemble variance. (c)~Raw and partial Spearman correlation of the score with molecular novelty and with local
label variance; controlling for variance largely removes the novelty association, not vice versa. Error bars: 95\%
bootstrap intervals over replicates.}
\label{fig:sel}
\end{figure}

In regime~A (Figure~\ref{fig:sel}; Appendix~\ref{app:e2}) the score tracks local label variance more than chemical
novelty: Spearman $\rho$ is \LipoRhoVar{} vs.\ \LipoRhoNov{} on lipophilicity and \BbbpRhoVar{} vs.\ \BbbpRhoNov{}
on BBBP (scaffold novelty: \LipoRhoScaf{} and \BbbpRhoScaf). Controlling for local label variance largely explains
away the novelty association---partial $\rho$ \LipoPartNovCI{} and \BbbpPartNovCI, reduced but still above
zero---whereas the variance association survives controlling for novelty (\LipoPartVar{} and \BbbpPartVar). A $10\%$
abstention budget is accordingly enriched for high-variance molecules (\LipoEnrVarTen$\times$ and
\BbbpEnrVarTen$\times$ over random) more than for novel ones (\LipoEnrNovTen$\times$ and \BbbpEnrNovTen$\times$). The
abstained set is also enriched for realized error (\LipoEnrErrTen$\times$ and \BbbpEnrErrTen$\times$), but error does
not follow the same pattern: on lipophilicity it is associated more with novelty ($\rho$ \LipoRhoErrNov) than with
local label variance (\LipoRhoErrVar), and the width--error association is only partly explained by local label
variance (partial $\rho$ \LipoPartWErrVar{} vs.\ raw \LipoRhoWErr), so on lipophilicity the score under-weights a
signal that matters for error. On BBBP the pattern reverses: error is associated with local label variance
(\BbbpRhoErrVar) far more than with novelty (\BbbpRhoErrNov).

Two caveats bound this interpretation. First, local label variance is not identified assay noise: variance among
distinct neighbours mixes measurement noise with structure--activity variation and activity cliffs
\citep{stumpfe2012activity,vantilborg2022exposing}, the data contain no replicate measurements \citep[cf.\ noise
ceilings in molecular datasets,][]{crusius2025fitting}, and the proxy
(replicate mean \NoiseLocalVar) exceeds the \eval{} mean squared error (\NoiseMse), so it cannot show that error is
noise-dominated (Appendix~\ref{app:noise}). Second, local label variance itself rises with novelty ($\rho$
\LipoRhoVarNov{} on lipophilicity), because novel molecules have less similar neighbours, so the partial correlations
cannot fully separate label heterogeneity from novelty-induced neighbourhood sparsity. We therefore read the score as
an estimate of local label dispersion in the training data rather than as a novelty estimate.

\subsection{Decision cost: uncertainty gates do not beat point-prediction ranking}
\label{sec:e3}
\begin{figure}[t]
\centering
\includegraphics[width=\linewidth]{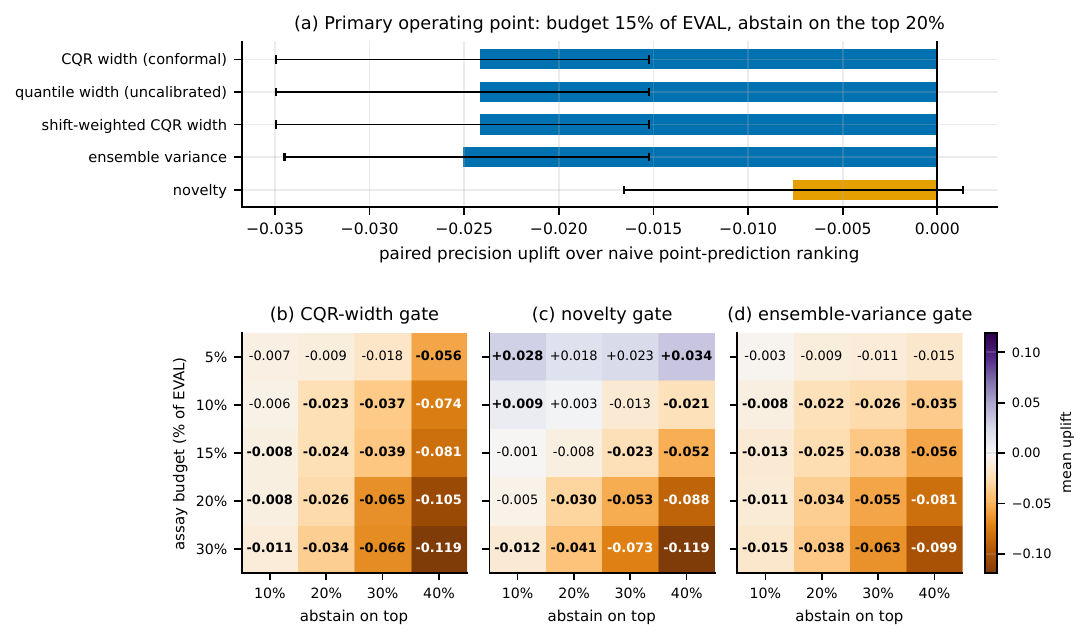}
\caption{\textbf{Uncertainty gates never beat point-prediction ranking; an exploratory novelty gate helps only at the
smallest budgets.} (a)~Paired precision uplift over \textsc{naive} at the primary operating point (95\% bootstrap
intervals; ranking by the conformal lower bound, \UpLowerBound, is omitted for scale). (b--d)~Mean paired uplift across
assay budgets (rows) and abstained fractions (columns) for the CQR-width, novelty and ensemble-variance gates; bold
values have 95\% intervals that exclude zero (not corrected for multiple comparisons).}
\label{fig:dec}
\end{figure}

At the primary operating point ($B{=}15\%$ of \eval, i.e.\ \Budget{} compounds; $\kappa{=}0.8$), ranking by point
prediction attains precision \PrecNaiveCI{} against a hit base rate of \BaseRate. No gate improves on it
(Table~\ref{tab:e3}, Figure~\ref{fig:dec}a). The paired uplift is \UpWidthCI{} for conformal width and, identically
by Remark~\ref{rem:width}, \UpWidthRawCI{} for the uncalibrated quantile width; \UpEnsVarCI{} for ensemble variance;
and \UpNoveltyCI{} for a novelty gate, which is indistinguishable from \textsc{naive} here. Ranking by the conformal
lower bound is far worse (\UpLowerBoundCI).

The exploratory sweep (Figure~\ref{fig:dec}b--d) covers \GateWidthCells{} settings per gate (\NBudgetSweep{} budgets
$\times$ \NKeep{} abstention fractions). Width and ensemble-variance gates are never significantly better than
\textsc{naive} and are significantly worse in \GateWidthSigNeg{} and \GateEnsSigNeg{} of the \GateWidthCells{}
settings, respectively. The novelty gate is significantly better in \GateNovSigPos{} settings (\NovSigPosFive{} at
$B{=}5\%$ and \NovSigPosTen{} at $B{=}10\%$; at $B{=}5\%$ its mean uplift ranges from $+$\NovFiveMin{} to
$+$\NovFiveMax{} across abstention fractions) and significantly worse in \GateNovSigNeg{} settings with larger budgets
or heavier abstention. With \GateTotalCells{} uncorrected comparisons and small effects, we treat this as a hypothesis.
The top of the point-prediction ranking suggests a mechanism: at $B{=}5\%$, the top-ranked molecules include
\TopNovFracFive{} from the top novelty quartile (vs.\ $25\%$ overall), and those novel molecules are hits less often
than the rest of the batch (precision \TopPrecNovFive{} vs.\ \TopPrecOtherFive), consistent with
\citet{sheridan2004similarity}; removing them helps only while the batch is small.

\begin{table}[t]
\centering
\caption{\textbf{Fixed-budget hit finding} on lipophilicity (regime~A, $B{=}15\%$ of \eval, gates drop the top $20\%$
by their score; mean and paired 95\% bootstrap interval over \NRep{} replicates; hit base rate \BaseRate). The three
width rows are identical by construction (Remark~\ref{rem:width}).}
\label{tab:e3}
\small
\input{generated/tab_e3}
\end{table}

\subsection{Where conformalization pays off: FDR-controlled selection}
\label{sec:fdr}
\begin{figure}[t]
\centering
\includegraphics[width=0.72\linewidth]{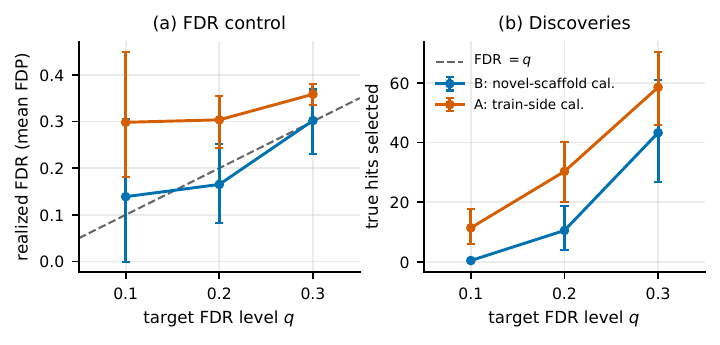}
\caption{\textbf{FDR control depends on calibration provenance.} (a)~Realized FDR of exact conformal selection
against the target level $q$; the dashed line is FDR $=q$. (b)~Mean number of true hits selected. 95\% bootstrap
intervals over replicates.}
\label{fig:fdr}
\end{figure}

Conformal selection offers a guarantee for the screening decision itself. Under novel-scaffold calibration (B), where
its assumptions hold, the realized FDR is statistically consistent with the target at every level
(Table~\ref{tab:fdr}, Figure~\ref{fig:fdr}): \JcBTenFdrCI, \JcBTwentyFdrCI{} and \JcBThirtyFdrCI{} at $q{=}0.1, 0.2,
0.3$ (at $q{=}0.1$, \JcBTenEmpty{} of replicates select nothing). Under training-scaffold calibration (A) it exceeds
the target at every level, with intervals excluding $q$: \JcATenFdrCI, \JcATwentyFdrCI{} and \JcAThirtyFdrCI. The
paired A$-$B difference is significant at \JcDiffSig{} of the three levels (\JcDiffTwentyCI{} at $q{=}0.2$). The
mechanism mirrors the coverage shortfall of \S\ref{sec:e1}: \cala{} molecules share scaffolds with the training set, so
the model is more accurate on them than on novel chemistry, too few calibration nulls receive high predictions, and
the $p$-values of novel test molecules come out too small.

Conformal selection contributes a stopping rule, not a better ranking. Because the count $N_j^{>}$ in its $p$-values is
monotone in the point prediction, the selected set overlaps heavily with the top-$k$ molecules by $\hat\mu$ (mean overlap
\JcBTwentyOverlap{} in~B at $q{=}0.2$; the rest reflects randomization among tied counts), and its pooled precision (\JcBTwentyPooled) is close to that of point-prediction
ranking at the same $k$ (\JcBTwentyPooledNaive). Its value is a data-driven batch size with an error guarantee---in~B
at $q{=}0.2$, \JcBTwentyN{} compounds on average (median \JcBTwentyNMed, maximum \JcBTwentyNMax)---and it requires
labels on \LipoNCalB{} novel-scaffold molecules. A more conservative variant we used initially
(Appendix~\ref{app:variant}) has a mean FDP under~A of \VarSelTwoFdpCI{} at $q{=}0.2$, consistent with the target,
at the cost of selecting nothing in \VarSelTwoEmpty{} of replicates.

\begin{table}[t]
\centering
\caption{\textbf{Exact conformal selection} \citep{jin2023selection} on lipophilicity with $p$-values from
Eq.~\eqref{eq:jc}. FDR is the replicate mean of FDP with FDP$\,:=0$ for empty selections (95\% bootstrap interval);
precision is pooled over all selected compounds; ``Top-$k$ by $\hat\mu$'' is the pooled precision of point-prediction
ranking at the same batch sizes; ``Overlap'' is the mean fraction of selected compounds among that top-$k$
(\NRep{} replicates).}
\label{tab:fdr}
\footnotesize
\input{generated/tab_fdr}
\end{table}

\section{Discussion}
\label{sec:disc}
\textbf{What conformal prediction buys here.} Conformalization provides guarantees---marginal coverage, FDR on a
selected batch---whose validity is a property of the calibration data, not of the model. Calibrating on held-out
molecules from the training scaffolds, the cheapest option available to a team, is not exchangeable with new
chemistry: on lipophilicity both nominal targets are missed there, coverage (\S\ref{sec:e1}) and FDR
(\S\ref{sec:fdr}), while BBBP coverage stays above nominal, though without a guarantee.
With calibration that is exchangeable with the screened chemistry, conformal selection yields a batch whose FDR is
consistent with the target, but the batch is small and variable, and its precision matches point-prediction ranking
at the same size. What conformalization does not do is decide what to abstain on: for split CQR that ranking is the
base model's, and here it follows local label dispersion more than novelty, while the only gate that ever improved hit
precision was an exploratory novelty gate at the smallest budgets.

\textbf{Reporting recommendations.} (1)~State the calibration regime alongside any coverage or FDR claim.
(2)~Distinguish the base model's uncertainty ranking from the conformal calibration step, and report the uncalibrated
control. (3)~For selection procedures, report the empty-selection frequency, the FDP convention, and point-prediction
precision at the same batch size. (4)~Do not attribute abstention to measurement noise without replicate
measurements. (5)~Evaluate abstention against point-prediction ranking at fixed budget, across budgets, and against a
plain novelty gate.

\textbf{Limitations.} We study one model family (gradient boosting on fingerprints) with a weak epistemic baseline
(three bagged members; stochastic-gradient-boosting ensembles are a stronger option \citep{malinin2021uncertainty});
tree ensembles may represent distance from the training data poorly, and a message-passing
network or deep ensemble \citep{yang2019analyzing,heid2024chemprop,lakshminarayanan2017simple} could behave
differently. All results are conditional on one deterministic scaffold split of two MoleculeNet endpoints; shift is
simulated rather than prospective \citep{sheridan2013time}, and \eval{} contains only singleton scaffolds. We do not
test other deployable calibration choices---a scaffold-held-out fold from the training pool, exact weighted conformal
prediction with an $x$-dependent $\hat Q(x)$ \citep{tibshirani2019conformal}, or weighted conformal selection
\citep{jin2026weighted}---which are the natural remedies for the regime-A failures. Local label variance is related to
novelty and only loosely bounds irreducible noise. The BBBP results include the non-empty-set rule. Decision and
selection experiments use the regression endpoint only, the sweep and the additional analyses are exploratory and
uncorrected for multiple comparisons, and $k$-nearest-neighbour ties make the local-variance statistics weakly
platform-dependent (Appendix~\ref{app:repro}).

\section{Conclusion}
Holding the model and evaluation molecules fixed and varying only the calibration fold separates what conformal
prediction guarantees from what it is often assumed to do. In our experiments with a gradient-boosted fingerprint
model under scaffold shift, coverage on the regression endpoint and FDR in conformal selection tracked calibration
provenance; the abstention ranking was the
base model's and followed local label variance more than novelty; and no uncertainty-based gate beat ranking by point
prediction at a fixed assay budget. Conformal guarantees were useful only when the calibration data resembled the
screened chemistry, which here required labels on novel-scaffold molecules.

\section*{Reproducibility statement}
All experiments are in a single notebook executed on Modal in a pinned environment (Python 3.13, RDKit 2026.03.6,
LightGBM 4.6.0, NumPy 2.1.3, pandas 2.2.3, scikit-learn 1.6.1, SciPy 1.16.3) with OpenMP threads fixed to two; tables,
figures and in-text numbers are generated from its outputs by a script, and cached per-molecule predictions are
retained so that follow-up analyses need no retraining (Appendix~\ref{app:repro}).
Code, configuration, results and cached predictions: \url{https://github.com/arnavsriva/conformal-abstention-scaffold-shift}.

\bibliographystyle{unsrtnat}
\bibliography{references}

\appendix
\section{Configuration and reproducibility}
\label{app:repro}
The hashed configuration (SHA-256 prefix \ConfigHash) (the replicate count was raised to \NRep{} after a three-replicate smoke test) fixes $\alpha{=}0.10$; \NRep{}
replicates; base seed $20260812$;
the pool fraction $0.70$ (largest scaffold groups first), \cala{} $0.15$ of the pool and \calb{} $0.34$ of the remainder;
ECFP radius 2 with 2048 bits; $k{=}10$ neighbours; abstention budgets $\{5,10,20,30\}\%$ with a top-quartile novelty
threshold; assay budget $0.15$; hit quantile $0.80$; FDR level $0.2$; and the $\alpha$ sweep $\{0.05,0.10,0.20\}$. Not
in the hash: the abstention fraction $\kappa{=}0.8$ and the LightGBM settings (400 trees, learning rate $0.05$, 31
leaves, min.\ child samples 10, subsample $0.8$, column subsample $0.6$; three members fit on bootstrap resamples,
since re-seeding alone gives near-zero ensemble variance), both fixed in the original code; and all additional analyses
(\S\ref{sec:setup}). In the $\alpha$ sweep, predictions are cached independently of $\alpha$, so the base quantile
models stay at levels $0.05/0.95$ and only the conformal level varies.

\textbf{Execution and determinism.} Models were trained once on Modal (2 OpenMP threads); every reported output comes
from a later Modal execution that re-ran all analysis cells from those cached predictions, with the final code. The
coverage, correlation, fixed-budget and initial-variant statistics also match an independent execution in a different
environment (Google Colab, two vCPUs) to three decimals; the additional analyses come from the Modal executions only, and
the diagnostics of Appendices~\ref{app:aps} and~\ref{app:noise} were computed once. Re-running the analysis cells on a
different machine exposes one platform dependence: the local-label-variance neighbours are chosen with an unstable sort
over Tanimoto similarities, which have many exact ties, so ties at the $k$-th neighbour are broken in a CPU-dependent way.
Between two machines, replicate-mean correlations differed by at most \TieRhoMax{} and enrichments by at most
\TieEnrMax; all other results computed with the same code on both machines were identical. A stable sort removes this
dependence.

\section{Correlation statistics}
\label{app:e2}
\begin{table}[H]
\centering
\caption{Spearman and partial Spearman (rank-residualized) correlations among the abstention score (width), the
per-molecule covariates and realized error in regime~A (mean and 95\% bootstrap interval over \NRep{} replicates). For
BBBP the score is effectively ensemble variance and the error is misclassification.}
\small
\input{generated/tab_e2}
\end{table}

\section{APS with and without the non-empty-set rule}
\label{app:aps}
Our APS implementation adds the top class to any prediction set that would otherwise be empty. Table~\ref{tab:aps}
compares coverage with and without this rule on BBBP, using identical randomization. Without it, coverage in the
exchangeable settings is close to nominal (random control \UnfRandCovCI, regime~B \UnfBCovCI), whereas regime~A
remains above nominal (\UnfACovCI) without a guarantee. Minority-class coverage without the rule is \UnfRandCovMin,
\UnfACovMin{} and \UnfBCovMin{} for the random control, A and B.
\begin{table}[H]
\centering
\caption{BBBP coverage at $\alpha{=}0.10$ with (``forced'') and without (``unforced'') the non-empty-set rule (mean over
\NRep{} replicates), the fraction of empty unforced sets, and the positive-class rate of \eval.}
\label{tab:aps}
\small
\input{generated/tab_aps}
\end{table}

\section{Class-conditional (Mondrian) calibration on BBBP}
\label{app:e4}
\begin{table}[H]
\centering
\caption{Marginal vs.\ class-conditional APS on BBBP (nominal 0.90; mean over \NRep{} replicates; includes the
non-empty-set rule). Class-conditional calibration raises minority coverage at the cost of larger sets. APS is
randomized: the regime-A marginal values coincide with Table~\ref{tab:e1}, while the regime-B values use different
random draws and differ from it in the third decimal.}
\small
\input{generated/tab_e4}
\end{table}

\section{Local label variance only loosely bounds irreducible noise}
\label{app:noise}
On lipophilicity, the \eval{} mean squared error is \NoiseMseCI{} and the mean local label variance over the $k{=}10$
nearest \fit{} neighbours is \NoiseLocalVarCI{} (replicate means; ratio of means \NoiseRatio, and the proxy exceeds the
MSE in every replicate, minimum ratio \NoiseRatioMin). Treating local label variance as an upper bound on irreducible
noise assumes that the neighbours' labels differ only through noise and smooth structure--activity variation; label
variance over \emph{distinct} molecules also reflects activity cliffs and reducible structure--activity gradients, and
the data are deduplicated by canonical SMILES, so no replicate measurements enter. The quantity is therefore a loose,
heuristic bound, and we interpret it as local label heterogeneity.

\section{Miscalibration area}
\label{app:miscal}
We initially reported a regression miscalibration area: the mean absolute difference between nominal and empirical
coverage over 19 levels $q\in\{0.05,\dots,0.95\}$ of Gaussian intervals $\hat\mu\pm\Phi^{-1}(0.5{+}q/2)\,\hat\sigma$,
with $\hat\sigma=|C(x)|/(2\Phi^{-1}(1-\alpha/2))$ derived from the conformal interval. On lipophilicity it is
\LipoMiscalRand{} (random), \LipoMiscalA{} (A) and \LipoMiscalB{} (B), ranking the valid regime~B worse than~A. The
metric assumes Gaussian intervals centred on $\hat\mu$, which CQR intervals are not, and scores levels other than the
target; we therefore report the conformal calibration curve across $\alpha$ (Figure~\ref{fig:cal}a) instead.

\section{The conformal-selection variant used initially}
\label{app:variant}
We initially used
$p_j=\big(1+\#\{i:\,y_i\le c,\ \hat\mu(x_i)\ge\hat\mu(x_j)\}\big)/(n_0+1)$, with $n_0$ the number of calibration
nulls, the null $y\le c$ and hits $y\ge c$. These $p$-values are pointwise no smaller than those of a Jin--Candès
construction with the same null, but the finite-sample FDR theorem of \citet{jin2023selection} is stated for the latter,
so we do not claim it for the variant. Under training-scaffold calibration at $q{=}0.2$ the variant selected nothing in
\VarSelTwoEmpty{} of replicates; its mean FDP (with FDP $:=0$ for empty selections) was \VarSelTwoFdpCI, and conditional
on a non-empty selection its precision was \VarSelTwoPrecNE{} and its FDP \VarSelTwoFdpNE{} over \VarSelTwoNNE{}
compounds; the unconditional precision, which averages empty selections in as zero, is \VarSelTwoPrec. Under novel-scaffold calibration the variant selected nothing in \VarBEmpty{} of replicates.

\end{document}

%% file: generated/numbers.tex
\newcommand{\BaseRate}{\ensuremath{0.218}}
\newcommand{\BbbpBrierA}{\ensuremath{0.065}}

\newcommand{\BbbpBrierRand}{\ensuremath{0.091}}

\newcommand{\BbbpCovACI}{\ensuremath{0.968}\ \ensuremath{[0.963,\allowbreak\,0.972]}}

\newcommand{\BbbpCovBCI}{\ensuremath{0.952}\ \ensuremath{[0.948,\allowbreak\,0.957]}}

\newcommand{\BbbpCovMinA}{\ensuremath{0.909}}

\newcommand{\BbbpCovMinB}{\ensuremath{0.866}}

\newcommand{\BbbpCovMinRand}{\ensuremath{0.787}}

\newcommand{\BbbpCovRandCI}{\ensuremath{0.930}\ \ensuremath{[0.924,\allowbreak\,0.937]}}

\newcommand{\BbbpEceA}{\ensuremath{0.046}}

\newcommand{\BbbpEceRand}{\ensuremath{0.075}}

\newcommand{\BbbpEnrErrTen}{\ensuremath{4.83}}

\newcommand{\BbbpEnrNovTen}{\ensuremath{1.33}}

\newcommand{\BbbpEnrVarTen}{\ensuremath{2.59}}

\newcommand{\BbbpNRem}{593}
\newcommand{\BbbpNllA}{\ensuremath{0.235}}

\newcommand{\BbbpNllRand}{\ensuremath{0.479}}

\newcommand{\BbbpPartNov}{\ensuremath{0.047}}
\newcommand{\BbbpPartNovCI}{\ensuremath{0.047}\ \ensuremath{[0.026,\allowbreak\,0.068]}}

\newcommand{\BbbpPartVar}{\ensuremath{0.497}}

\newcommand{\BbbpPos}{\ensuremath{0.76}}

\newcommand{\BbbpRhoErrNov}{\ensuremath{0.041}}

\newcommand{\BbbpRhoErrVar}{\ensuremath{0.263}}

\newcommand{\BbbpRhoNov}{\ensuremath{0.158}}

\newcommand{\BbbpRhoScaf}{\ensuremath{0.190}}

\newcommand{\BbbpRhoVar}{\ensuremath{0.528}}

\newcommand{\BbbpRhoWidthEns}{\ensuremath{0.982}}

\newcommand{\BbbpSizeOne}{\ensuremath{83\%}}
\newcommand{\Budget}{124}
\newcommand{\ConfigHash}{\texttt{926171a32b97}}
\newcommand{\EvPosA}{\ensuremath{0.82}}

\newcommand{\EvPosRand}{\ensuremath{0.76}}

\newcommand{\GateEnsSigNeg}{16}

\newcommand{\GateNovSigNeg}{10}
\newcommand{\GateNovSigPos}{3}
\newcommand{\GateTotalCells}{60}
\newcommand{\GateWidthCells}{20}

\newcommand{\GateWidthSigNeg}{16}

\newcommand{\HitTieFrac}{\ensuremath{0.2\%}}

\newcommand{\JcATenFdrCI}{\ensuremath{0.298}\ \ensuremath{[0.180,\allowbreak\,0.448]}}

\newcommand{\JcAThirtyFdrCI}{\ensuremath{0.358}\ \ensuremath{[0.335,\allowbreak\,0.380]}}

\newcommand{\JcATwentyFdr}{\ensuremath{0.303}}
\newcommand{\JcATwentyFdrCI}{\ensuremath{0.303}\ \ensuremath{[0.243,\allowbreak\,0.355]}}

\newcommand{\JcBTenEmpty}{\ensuremath{67\%}}

\newcommand{\JcBTenFdrCI}{\ensuremath{0.139}\ \ensuremath{[0.000,\allowbreak\,0.306]}}

\newcommand{\JcBThirtyFdrCI}{\ensuremath{0.302}\ \ensuremath{[0.230,\allowbreak\,0.368]}}

\newcommand{\JcBTwentyFdrCI}{\ensuremath{0.165}\ \ensuremath{[0.082,\allowbreak\,0.252]}}

\newcommand{\JcBTwentyN}{\ensuremath{15.9}}
\newcommand{\JcBTwentyNMax}{\ensuremath{80}}
\newcommand{\JcBTwentyNMed}{\ensuremath{5.5}}
\newcommand{\JcBTwentyOverlap}{\ensuremath{0.83}}
\newcommand{\JcBTwentyPooled}{\ensuremath{0.661}}
\newcommand{\JcBTwentyPooledNaive}{\ensuremath{0.657}}

\newcommand{\JcDiffSig}{1}

\newcommand{\JcDiffTwentyCI}{\ensuremath{0.138}\ \ensuremath{[0.025,\allowbreak\,0.245]}}

\newcommand{\LipoEnrErrTen}{\ensuremath{1.65}}

\newcommand{\LipoEnrNovTen}{\ensuremath{1.25}}

\newcommand{\LipoEnrVarTen}{\ensuremath{1.73}}

\newcommand{\LipoEvalJaccard}{\ensuremath{0.49}}
\newcommand{\LipoMiscalA}{\ensuremath{0.034}}

\newcommand{\LipoMiscalB}{\ensuremath{0.081}}

\newcommand{\LipoMiscalRand}{\ensuremath{0.119}}

\newcommand{\LipoNCalB}{428}
\newcommand{\LipoNEval}{832}

\newcommand{\LipoNRem}{1260}
\newcommand{\LipoPartNov}{\ensuremath{0.067}}
\newcommand{\LipoPartNovCI}{\ensuremath{0.067}\ \ensuremath{[0.055,\allowbreak\,0.077]}}

\newcommand{\LipoPartVar}{\ensuremath{0.254}}

\newcommand{\LipoPartWErrVar}{\ensuremath{0.113}}

\newcommand{\LipoPicpA}{\ensuremath{0.841}}
\newcommand{\LipoPicpACI}{\ensuremath{0.841}\ \ensuremath{[0.831,\allowbreak\,0.849]}}

\newcommand{\LipoPicpB}{\ensuremath{0.904}}
\newcommand{\LipoPicpBCI}{\ensuremath{0.904}\ \ensuremath{[0.894,\allowbreak\,0.913]}}

\newcommand{\LipoPicpRandCI}{\ensuremath{0.905}\ \ensuremath{[0.897,\allowbreak\,0.912]}}

\newcommand{\LipoRhoErrNov}{\ensuremath{0.220}}

\newcommand{\LipoRhoErrVar}{\ensuremath{0.156}}

\newcommand{\LipoRhoNov}{\ensuremath{0.184}}

\newcommand{\LipoRhoScaf}{\ensuremath{0.165}}

\newcommand{\LipoRhoVar}{\ensuremath{0.309}}

\newcommand{\LipoRhoVarNov}{\ensuremath{0.426}}

\newcommand{\LipoRhoWErr}{\ensuremath{0.157}}

\newcommand{\LipoSd}{\ensuremath{1.20}}

\newcommand{\MinDiffABCI}{\ensuremath{0.043}\ \ensuremath{[0.025,\allowbreak\,0.061]}}

\newcommand{\MonBMargSet}{\ensuremath{1.10}}

\newcommand{\MonBMondCovMin}{\ensuremath{0.942}}
\newcommand{\MonBMondSet}{\ensuremath{1.19}}
\newcommand{\NBbbp}{1975}
\newcommand{\NBudgetSweep}{5}
\newcommand{\NKeep}{4}
\newcommand{\NLipo}{4200}
\newcommand{\NRep}{18}
\newcommand{\NRepSweep}{9}
\newcommand{\NoiseLocalVar}{\ensuremath{0.856}}
\newcommand{\NoiseLocalVarCI}{\ensuremath{0.856}\ \ensuremath{[0.848,\allowbreak\,0.863]}}

\newcommand{\NoiseMse}{\ensuremath{0.615}}
\newcommand{\NoiseMseCI}{\ensuremath{0.615}\ \ensuremath{[0.601,\allowbreak\,0.628]}}

\newcommand{\NoiseRatio}{\ensuremath{1.39}}
\newcommand{\NoiseRatioMin}{\ensuremath{1.31}}
\newcommand{\NovFiveMax}{\ensuremath{0.034}}
\newcommand{\NovFiveMin}{\ensuremath{0.018}}

\newcommand{\NovSigPosFive}{2}
\newcommand{\NovSigPosTen}{1}

\newcommand{\OverlapFitEval}{0}

\newcommand{\PrecNaiveCI}{\ensuremath{0.615}\ \ensuremath{[0.601,\allowbreak\,0.628]}}

\newcommand{\RuleShareA}{\ensuremath{34\%}}
\newcommand{\RuleShareB}{\ensuremath{97\%}}
\newcommand{\RuleShareRand}{\ensuremath{81\%}}
\newcommand{\SweepAFive}{\ensuremath{0.905}}

\newcommand{\SweepATen}{\ensuremath{0.836}}

\newcommand{\SweepATwenty}{\ensuremath{0.736}}

\newcommand{\SweepBFive}{\ensuremath{0.952}}

\newcommand{\SweepBTen}{\ensuremath{0.902}}

\newcommand{\SweepBTwenty}{\ensuremath{0.804}}

\newcommand{\SzCalA}{441}

\newcommand{\SzFit}{2499}

\newcommand{\TieEnrMax}{\ensuremath{0.020}}
\newcommand{\TieRhoMax}{\ensuremath{4.4\times10^{-4}}}

\newcommand{\TopNovFracFive}{\ensuremath{22\%}}

\newcommand{\TopPrecNovFive}{\ensuremath{0.579}}

\newcommand{\TopPrecOtherFive}{\ensuremath{0.732}}

\newcommand{\UnfACov}{\ensuremath{0.945}}
\newcommand{\UnfACovCI}{\ensuremath{0.945}\ \ensuremath{[0.934,\allowbreak\,0.956]}}

\newcommand{\UnfACovMin}{\ensuremath{0.891}}

\newcommand{\UnfBCov}{\ensuremath{0.901}}
\newcommand{\UnfBCovCI}{\ensuremath{0.901}\ \ensuremath{[0.889,\allowbreak\,0.914]}}

\newcommand{\UnfBCovMin}{\ensuremath{0.834}}

\newcommand{\UnfRandCov}{\ensuremath{0.906}}
\newcommand{\UnfRandCovCI}{\ensuremath{0.906}\ \ensuremath{[0.894,\allowbreak\,0.917]}}

\newcommand{\UnfRandCovMin}{\ensuremath{0.771}}

\newcommand{\UpEnsVarCI}{\ensuremath{-0.025}\ \ensuremath{[-0.034,\allowbreak\,-0.015]}}

\newcommand{\UpLowerBound}{\ensuremath{-0.171}}
\newcommand{\UpLowerBoundCI}{\ensuremath{-0.171}\ \ensuremath{[-0.185,\allowbreak\,-0.156]}}

\newcommand{\UpNoveltyCI}{\ensuremath{-0.008}\ \ensuremath{[-0.017,\allowbreak\,0.001]}}

\newcommand{\UpWidthCI}{\ensuremath{-0.024}\ \ensuremath{[-0.035,\allowbreak\,-0.015]}}

\newcommand{\UpWidthRawCI}{\ensuremath{-0.024}\ \ensuremath{[-0.035,\allowbreak\,-0.015]}}

\newcommand{\VarBEmpty}{\ensuremath{94\%}}

\newcommand{\VarSelTwoEmpty}{\ensuremath{39\%}}

\newcommand{\VarSelTwoFdpCI}{\ensuremath{0.186}\ \ensuremath{[0.115,\allowbreak\,0.257]}}

\newcommand{\VarSelTwoFdpNE}{\ensuremath{0.305}}

\newcommand{\VarSelTwoNNE}{\ensuremath{45.9}}
\newcommand{\VarSelTwoPrec}{\ensuremath{0.425}}
\newcommand{\VarSelTwoPrecNE}{\ensuremath{0.695}}
\newcommand{\WidthIdentityMaxAbs}{\ensuremath{0}}

%% file: generated/tab_e1.tex
\begin{tabular}{@{}lccc@{}}
\toprule
\multicolumn{4}{@{}l}{\textit{Lipophilicity (regression; CQR)}}\\
\midrule
Split / calibration & PICP & MPIW & RMSE \\
\midrule
random (control) & ${0.905}$ {\scriptsize[0.897, 0.912]} & $2.81$ & $0.658$ \\
scaffold, A (train-side) & $\mathbf{0.841}$ {\scriptsize[0.831, 0.849]} & $2.67$ & $0.784$ \\
scaffold, B (novel-scaffold) & ${0.904}$ {\scriptsize[0.894, 0.913]} & $3.08$ & $0.784$ \\
\bottomrule
\end{tabular}

\vspace{6pt}
\begin{tabular}{@{}lcccccc@{}}
\toprule
\multicolumn{7}{@{}l}{\textit{BBBP (classification; APS)}}\\
\midrule
Split / calibration & Coverage & Cov.\ (minority) & $|\mathcal{C}|$ & ECE & Brier & NLL \\
\midrule
random (control) & $0.930$ {\scriptsize[0.924, 0.937]} & $0.787$ & $1.10$ & $0.075$ & $0.091$ & $0.479$ \\
scaffold, A (train-side) & $0.968$ {\scriptsize[0.963, 0.972]} & $0.909$ & $1.17$ & $0.046$ & $0.065$ & $0.235$ \\
scaffold, B (novel-scaffold) & $0.952$ {\scriptsize[0.948, 0.957]} & $0.866$ & $1.10$ & $0.046$ & $0.065$ & $0.235$ \\
\bottomrule
\end{tabular}

%% file: generated/tab_e3.tex
\begin{tabular}{@{}lcc@{}}
\toprule
Policy (select top-$B$ by point prediction among retained) & Precision & Paired uplift vs.\ \textsc{naive} \\
\midrule
point prediction, no UQ (\textsc{naive}) & $0.615$ {\scriptsize[0.601, 0.628]} & --- \\
abstain: CQR width (conformal) & $0.591$ {\scriptsize[0.575, 0.607]} & $-0.024$ {\scriptsize[-0.035, -0.015]} \\
abstain: quantile width (uncalibrated) & $0.591$ {\scriptsize[0.575, 0.607]} & $-0.024$ {\scriptsize[-0.035, -0.015]} \\
abstain: shift-weighted CQR width & $0.591$ {\scriptsize[0.575, 0.607]} & $-0.024$ {\scriptsize[-0.035, -0.015]} \\
abstain: ensemble variance & $0.590$ {\scriptsize[0.577, 0.603]} & $-0.025$ {\scriptsize[-0.034, -0.015]} \\
abstain: novelty ($1-\max$ Tanimoto) & $0.607$ {\scriptsize[0.593, 0.622]} & $-0.008$ {\scriptsize[-0.017, +0.001]} \\
rank by conformal lower bound & $0.444$ {\scriptsize[0.432, 0.457]} & $-0.171$ {\scriptsize[-0.185, -0.156]} \\
\bottomrule
\end{tabular}

%% file: generated/tab_fdr.tex
\begin{tabular}{@{}lccccccc@{}}
\toprule
Calibration & $q$ & FDR (mean FDP) & Empty & \# sel.\ mean (median) & Prec.\ (pooled) & Top-$k$ by $\hat\mu$ & Overlap \\
\midrule
A (train-side) & $0.1$ & $0.298$ {\scriptsize[0.180, 0.448]} & $28\%$ & $16.1$ ($12.5$) & $0.707$ & $0.714$ & $0.71$ \\
A (train-side) & $0.2$ & $0.303$ {\scriptsize[0.243, 0.355]} & $6\%$ & $44.8$ ($46.5$) & $0.677$ & $0.677$ & $0.89$ \\
A (train-side) & $0.3$ & $0.358$ {\scriptsize[0.335, 0.380]} & $0\%$ & $91.8$ ($91.0$) & $0.639$ & $0.639$ & $0.99$ \\
\midrule
B (novel-scaffold) & $0.1$ & $0.139$ {\scriptsize[0.000, 0.306]} & $67\%$ & $0.7$ ($0.0$) & $0.538$ & $0.769$ & $0.28$ \\
B (novel-scaffold) & $0.2$ & $0.165$ {\scriptsize[0.082, 0.252]} & $33\%$ & $15.9$ ($5.5$) & $0.661$ & $0.657$ & $0.83$ \\
B (novel-scaffold) & $0.3$ & $0.302$ {\scriptsize[0.230, 0.368]} & $17\%$ & $72.2$ ($61.0$) & $0.600$ & $0.603$ & $1.00$ \\
\bottomrule
\end{tabular}

%% file: generated/tab_e2.tex
\begin{tabular}{@{}lcc@{}}
\toprule
Statistic (regime A) & Lipophilicity & BBBP \\
\midrule
$\rho$(width, molecular novelty) & $+0.184$ {\scriptsize[+0.173, +0.194]} & $+0.158$ {\scriptsize[+0.139, +0.176]} \\
$\rho$(width, scaffold novelty) & $+0.165$ {\scriptsize[+0.154, +0.175]} & $+0.190$ {\scriptsize[+0.174, +0.206]} \\
$\rho$(width, local label variance) & $+0.309$ {\scriptsize[+0.295, +0.323]} & $+0.528$ {\scriptsize[+0.514, +0.543]} \\
$\rho$(width, $|y-\hat y|$ or error) & $+0.157$ {\scriptsize[+0.147, +0.167]} & $+0.377$ {\scriptsize[+0.363, +0.390]} \\
partial $\rho$(width, novelty $\mid$ variance) & $+0.067$ {\scriptsize[+0.055, +0.077]} & $+0.047$ {\scriptsize[+0.026, +0.068]} \\
partial $\rho$(width, variance $\mid$ novelty) & $+0.254$ {\scriptsize[+0.238, +0.270]} & $+0.497$ {\scriptsize[+0.483, +0.512]} \\
$\rho$(error, molecular novelty) & $+0.220$ {\scriptsize[+0.209, +0.233]} & $+0.041$ {\scriptsize[+0.018, +0.065]} \\
$\rho$(error, local label variance) & $+0.156$ {\scriptsize[+0.145, +0.167]} & $+0.263$ {\scriptsize[+0.244, +0.281]} \\
$\rho$(local label variance, novelty) & $+0.426$ {\scriptsize[+0.415, +0.437]} & $+0.218$ {\scriptsize[+0.205, +0.229]} \\
partial $\rho$(width, error $\mid$ variance) & $+0.113$ {\scriptsize[+0.103, +0.121]} & $+0.156$ {\scriptsize[+0.143, +0.170]} \\
partial $\rho$(width, error $\mid$ novelty) & $+0.121$ {\scriptsize[+0.112, +0.130]} & $+0.183$ {\scriptsize[+0.173, +0.194]} \\
\bottomrule
\end{tabular}

%% file: generated/tab_aps.tex
\begin{tabular}{@{}lcccccc@{}}
\toprule
 & \multicolumn{2}{c}{Coverage} & \multicolumn{2}{c}{Cov.\ (minority)} & Empty sets & \eval{} positive \\
Split / calibration & forced & unforced & forced & unforced & (unforced) & rate \\
\midrule
random (control) & $0.930$ & $0.906$ & $0.787$ & $0.771$ & $2.6\%$ & $0.76$ \\
scaffold, A (train-side) & $0.968$ & $0.945$ & $0.909$ & $0.891$ & $2.3\%$ & $0.82$ \\
scaffold, B (novel-scaffold) & $0.952$ & $0.901$ & $0.866$ & $0.834$ & $5.3\%$ & $0.82$ \\
\bottomrule
\end{tabular}

%% file: generated/tab_e4.tex
\begin{tabular}{@{}llcccc@{}}
\toprule
Calibration & Quantile & Coverage & Cov.\ (minority) & Cov.\ (majority) & $|\mathcal{C}|$ \\
\midrule
A (train-side) & marginal & $0.968$ & $0.909$ & $0.981$ & $1.17$ \\
A (train-side) & class-conditional & $0.961$ & $0.969$ & $0.960$ & $1.34$ \\
B (novel-scaffold) & marginal & $0.951$ & $0.865$ & $0.970$ & $1.10$ \\
B (novel-scaffold) & class-conditional & $0.962$ & $0.942$ & $0.967$ & $1.19$ \\
\bottomrule
\end{tabular}